\documentclass[10pt,twocolumn,letterpaper]{article}

\usepackage{cvpr}
\usepackage{times}
\usepackage{epsfig}
\usepackage{graphicx}
\usepackage{amsmath}
\usepackage{amssymb}

\usepackage[pagebackref=true,breaklinks=true,letterpaper=true,colorlinks,bookmarks=false]{hyperref}

\cvprfinalcopy 

\def\cvprPaperID{****} 

\begin{document}

\title{The Deepfake Detection Challenge (DFDC) Preview Dataset}
\author{Brian Dolhansky, Russ Howes, Ben Pflaum, Nicole Baram, Cristian Canton Ferrer\\
AI Red Team, Facebook AI\\
}

\maketitle

\begin{abstract}
In this paper, we introduce a preview of the Deepfakes Detection Challenge (DFDC) dataset consisting of 5K videos featuring two facial modification algorithms. A data collection campaign has been carried out where participating actors have entered into an agreement to the use and manipulation of their likenesses in our creation of the dataset. Diversity in several axes (gender, skin-tone, age, etc.) has been considered and actors recorded videos with arbitrary backgrounds thus bringing visual variability. Finally, a set of specific metrics to evaluate the performance have been defined and two existing models for detecting deepfakes have been tested to provide a reference performance baseline. The DFDC dataset preview can be downloaded at: 
\begin{center}\href{https://deepfakedetectionchallenge.ai/}{deepfakedetectionchallenge.ai}\end{center}
\end{abstract}

\section{Introduction}


\begin{figure*}
\begin{center}
    \includegraphics[width=0.2\textwidth]{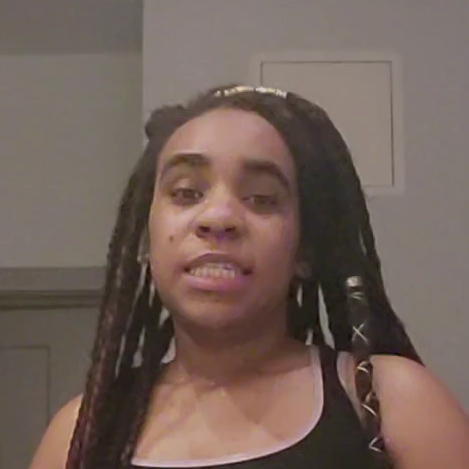}\includegraphics[width=0.2\textwidth]{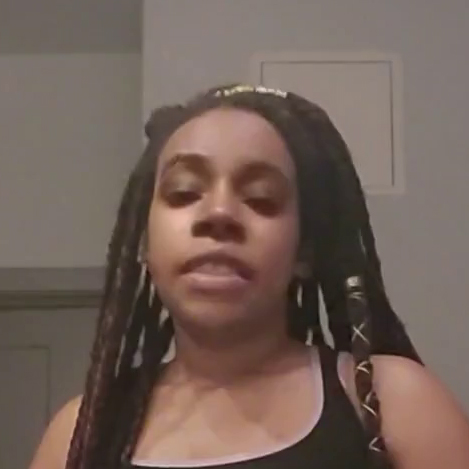}\hspace{1mm}\includegraphics[width=0.2\textwidth]{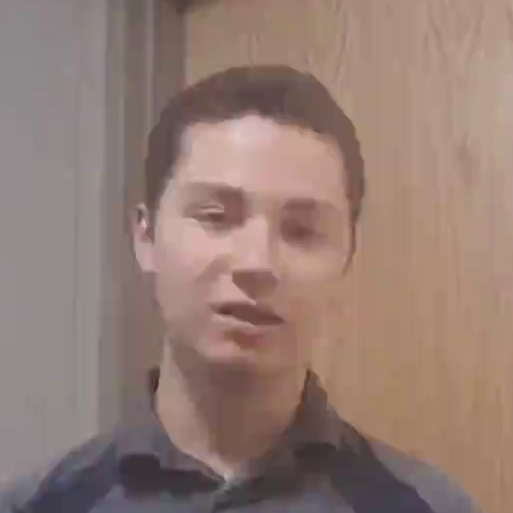}\includegraphics[width=0.2\textwidth]{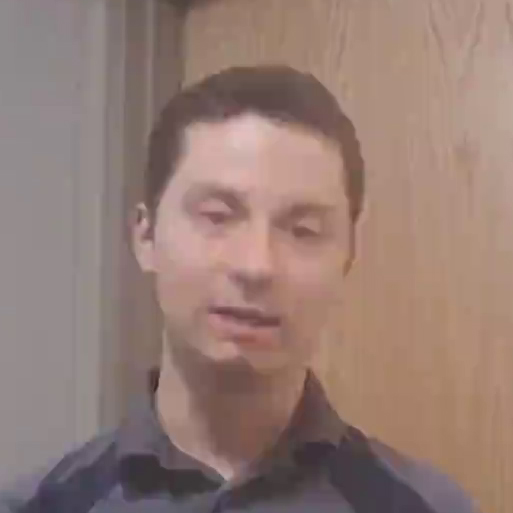}\\
    \vspace*{1mm}
    \includegraphics[width=0.2\textwidth]{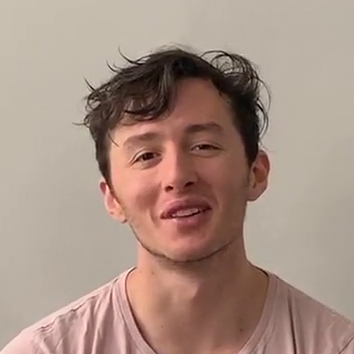}\includegraphics[width=0.2\textwidth]{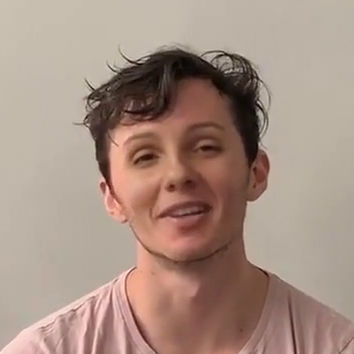}\hspace{1mm}\includegraphics[width=0.2\textwidth]{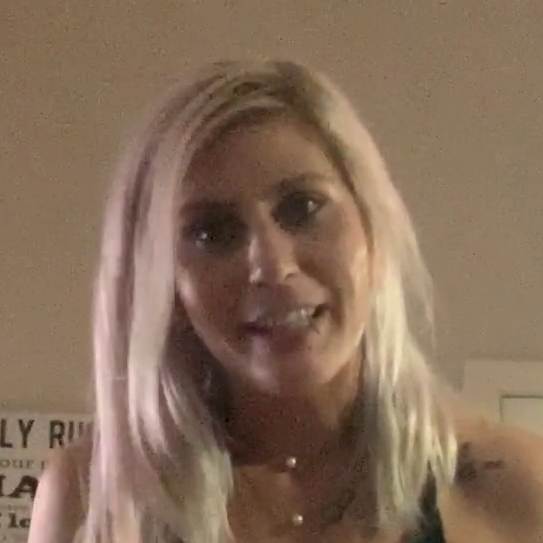}\includegraphics[width=0.2\textwidth]{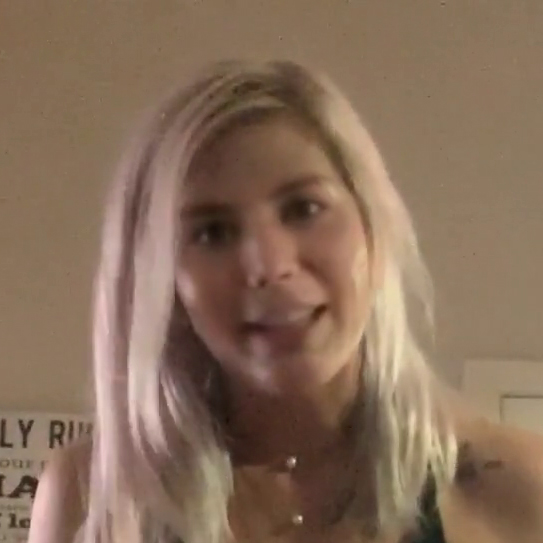}\\
    \vspace*{1mm}
    \includegraphics[width=0.2\textwidth]{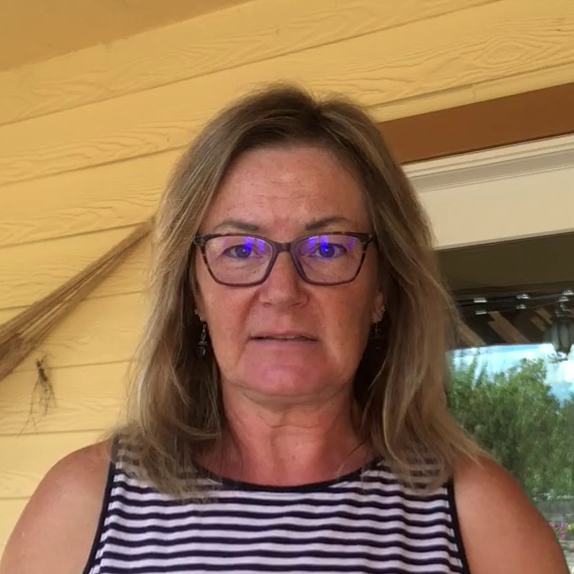}\includegraphics[width=0.2\textwidth]{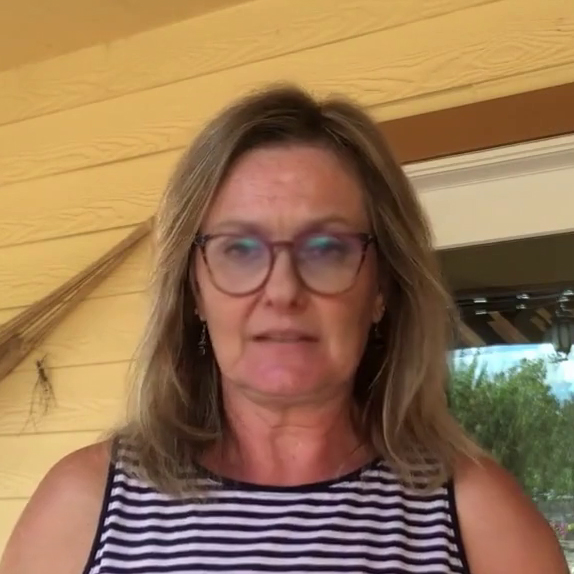}\hspace{1mm}\includegraphics[width=0.2\textwidth]{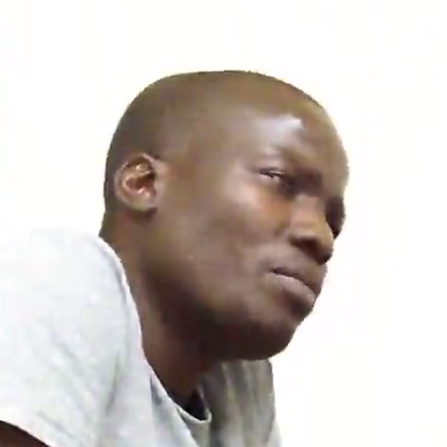}\includegraphics[width=0.2\textwidth]{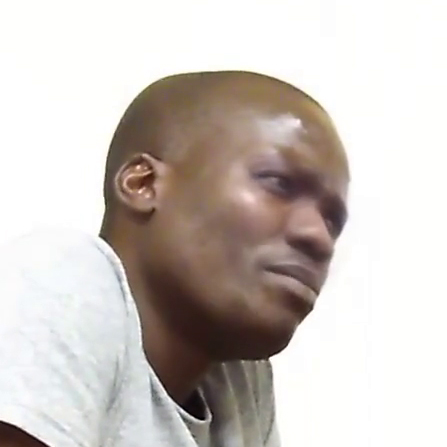}\\
    \vspace*{1mm}
    \includegraphics[width=0.2\textwidth]{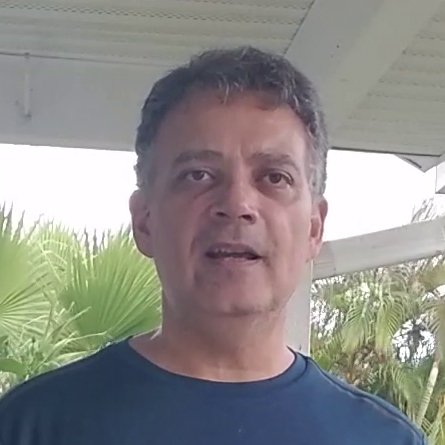}\includegraphics[width=0.2\textwidth]{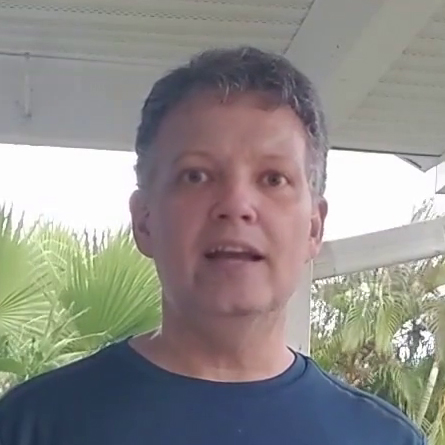}\hspace{1mm}\includegraphics[width=0.2\textwidth]{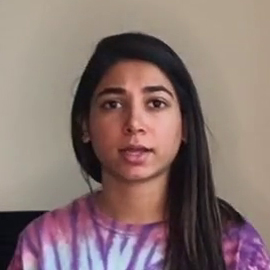}\includegraphics[width=0.2\textwidth]{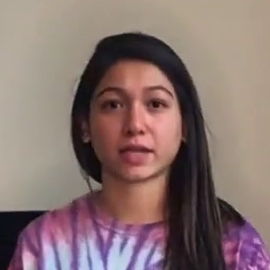}\\
\end{center}
\caption{Some example face swaps from the dataset.}
\vspace*{-2mm}
\end{figure*}

\par Detection of manipulated visual content is a research focus of intense interest in academia and industry, and a pressing topic of conversation in broader society. One type of manipulated visual content, high-quality videos containing facial manipulations (colloquially referred to as 'deepfakes'), has become particularly salient over the last two years. While there are benign and even humorous applications of deepfake videos, when created with malicious intent they have the potential to intimidate and harass individuals \cite{floridi2018}, or propagate disinformation that can adversely impact elections or financial systems \cite{chesney2018,barrett2019}. 

\par The umbrella term 'deepfakes' refers to a wide variety of methods for face swapping and manipulation, including methods that use state-of-the-art techniques from computer vision and deep learning, as well as other, simpler methods. The proliferation of ways to generate deepfakes has in turn yielded research to accurately detect videos with facial manipulations. Many of these detection models are based on modern machine learning techniques, and are improved with the use of large quantities of robust training data representing multiple generation methods.

\par A number of datasets featuring manipulated facial footage have been recently published with the purpose of training and evaluating facial manipulation detection and classification. These include the Celeb-DF \cite{li2019celebdf}, UADFV \cite{yang2019}, DeepFake-TIMIT \cite{korshunov2018}, VTD dataset \cite{al-sanjary2018}, FaceForensics \cite{roessler2018faceforensics} and its successor FaceForensics++ \cite{roessler2019faceforensics}. Table \ref{Table:Datasets} summarized the specs of the most relevant datasets in the deepfake space. The DFDC dataset aims to improve upon this work via: data diversity, agreement from persons depicted in the dataset, size of dataset, and metrics, among others.

\par The Deepfake Detection Challenge (DFDC) was announced in September 2019 \cite{dfdc2019} as a joint effort between industry, academia, and civil society organizations to invigorate the research related to the detection of facial manipulation. An integral part of the challenge is a dataset consisting of a large quantity of videos containing human faces, with accompanying labels describing whether they were generated using facial manipulation techniques. All videos in the dataset are created by entering into agreements from paid actors, and the dataset will be made freely available to the community for the development, testing, and analysis of techniques for detecting videos with manipulated faces. Developers interested to participate in this challenge will have to request access and agree with the terms of service of the DFDC. 

\par As part of the development process for the DFDC dataset, we introduce in this paper a preview dataset consisting of around 5000 videos (original and manipulated). We describe the properties of the starter dataset and share relevant benchmark results using existing deepfake detection methods.


\section{Dataset Construction}

\par In order to gather the necessary footage to build the DFDC dataset, a data collection campaign has been devised where actors were crowdsourced ensuring a variability in gender, skin tone and age. Each participant submitted a set of videos where they answered a set of pre-specified questions. The videos include varied lighting conditions and head poses, and participants were able to record their videos with any background they desired, which yielded visually diverse backgrounds. One key differentiating factor from other existing datasets is that is that actors have agreed to participate in the creation of the dataset which uses and modifies their likeness. The rough approximation of the general distribution of gender and race across this preview dataset is 74\% female and 26\% male; and 68\% Caucasian, 20\% African-American, 9\% east-Asian, and 3\% south-Asian. As we continue our crowdsourced data capture campaign, we will keep working on improving the diversity towards the publication of the final DFDC dataset. No publicly available data or data from social media sites was used to create this dataset.

\par For this first version of the DFDC dataset, a small set of 66 individuals where chosen from the pool of crowdsourced actors, and split into a training and a testing set. This was done to avoid cross-set face swaps. Two methods were selected to generate face swaps (noted as methods A and B in the dataset); with the intention of representing the real adversarial space of facial manipulation, no further details of the employed methods are disclosed to the participants. A number of state-of-the-art methods will be applied to generate these videos, exploring the whole gamut of manipulation techniques available to generate such tamperings.

\begin{table*}[!t]
\begin{center}
\begin{tabular}{|l c c c c|} 
\hline
\multicolumn{1}{|c}{Dataset} & \multicolumn{1}{c}{\begin{tabular}[c]{@{}c@{}}Ratio\\ tampered:original\end{tabular}} & Total videos & Source & Participants Consent\\ [0.5ex] 
\hline\hline
Celeb-DF \cite{li2019celebdf} & $1:0.51$ & $1203$ & YouTube & N \\ 
\hline
FaceForensics \cite{roessler2018faceforensics} & $1:1.00$ & $2008$ & YouTube & N\\ 
\hline
FaceForensics++ \cite{roessler2019faceforensics} & $1:0.25$ & $5000$ & YouTube & N \\ 
\hline
\multicolumn{1}{|l}{\begin{tabular}[l]{@{}l@{}}DeepFakeDetection \cite{dufour_gully_2019}\\(part of FaceForensics++)\end{tabular}} & $1:0.12$ & $3363$ & Actors & Y \\ 
\hline
{\bf DFDC Preview Dataset} & $1:0.28$ & $5214$ & Actors & Y \\ 
\hline
\end{tabular}
\end{center}
\caption{Specs of the most relevant deepfake datasets in the literature.}
\label{Table:Datasets}
\end{table*}

\par A number of face swaps were computed across subjects with similar appearances, where each appearance was inferred from facial attributes (skin tone, facial hair, glasses, etc.). After a given pairwise model was trained on two identities, we swapped each identity onto the other's videos. Hereafter, we refer to the identity in the base video as the ``target'' identity, and the identity of the face swapped onto the video as the ``swapped'' identity. In this preview DFDC dataset, all base and target videos are provided as part of the training corpus.


 \par The initial algorithm  in the dataset used to produce this dataset (method\_A) does not produce sharp or believable face swaps if the subject's face is too close to the camera, so selfie videos or other close-up recordings resulted in easy-to-spot fakes. Therefore, we initially filtered all videos by their average face size ratio, which measures the ratio of the maximum dimension of the face bounding boxes from a sampled set of frames to the minimum dimension of the video. All swaps were performed on videos where this measure was less than 0.25; in order to enrich the original training set, we included all original videos of the identities in this dataset where this measure was less than 0.3 (regardless of whether or not the video appeared in a swap). We additionally provide a second method (method\_B) that generally produces lower-quality swaps, but is similar to other off-the-shelf face swap algorithms.
 
 \par After some rough filtering, for each original or swapped video, we removed the first five seconds of the video as subjects were often seen setting up their cameras during this time. From the remaining length of the video, we extracted multiple 15 second clips - generally three clips per video if the length was over 50 seconds. All of the clips that comprised the training set were left at their original resolution and quality, so deriving appropriate augmentations of the training set is left as an exercise to the researcher. However, for each target video in the test set, we randomly selected two clips out of three and applied augmentations that approximate actual degradations seen in real-life video distributions. Specifically, these augmentations were (1) reduce the FPS of the video to 15; (2) reduce the resolution of the video to 1/4 of its original size; and (3) reduce the overall encoding quality. In this dataset, no video was subjected to more than one augmentation. The third remaining test clip did not undergo any augmentation. After adding these original video clips and their augmentations, we end up with a total of 4,464 unique training clips and 780 unique test clips.


 \par All information regarding the swapped and target identities for each video, along with the train or test set assignment and any augmentations applied to a video are listed in the file \texttt{dataset.json}, located in the dataset root directory. Swapped video filenames contain two IDs - the first is the swapped ID, and the second is the target ID. The final identifiers in the video refer to which target video the swap was produced from, and the clip within the target video that was used. Finally, the dataset can be downloaded at:\\
 \vspace*{-3mm}
 \begin{center}\href{https://deepfakedetectionchallenge.ai/}{deepfakedetectionchallenge.ai}\end{center}
\section{Evaluation metrics}

\par All relevant datasets addressing the task of Deepfake detection (see Table \ref{Table:Datasets}) produce metrics that are strongly influenced by the distribution of positive and negative examples present in the test set. As a result, it is difficult to quantitatively measure how any of the methods evaluated on those datasets would perform when facing the real production traffic that any social media company would ingest. This is particularly relevant when evaluating the impact of false positives (FP) and their associated actions hence the need to capture this effect into the evaluation metrics for this competition.


\par Current prevalence of deepfakes (compared to unaltered videos) in \emph{organic traffic} is much lower than that corresponding to the ratios for the datasets in Table \ref{Table:Datasets}. If we assume that the ratio between deepfake and unaltered videos is $1:x$ in organic traffic and $1:y$ in a deepfakes dataset, it is likely that $x\gg y$. Although it is not practical to construct a dataset that mimics the statistics of organic traffic, it is critical to define metrics that capture these differences. We can define a \emph{weighted precision} for a deepfakes dataset as a very rough approximation of the precision that would be computed by evaluating on a dataset representative of organic traffic. Assuming the ratios of unaltered to tampered videos differ between a test dataset and organic traffic by a factor of $\alpha = x/y$, we define weighted precision wP and (standard) recall R as
\begin{equation}
\textrm{wP} = \frac{\textrm{TP}}{\textrm{TP}+\alpha\textrm{FP}},\;\;\;\;\;\; \textrm{R} = \frac{\textrm{TP}}{\textrm{TP}+\textrm{FN}}, 
\end{equation}
where TP, FP, and FN signify true positives, false positives, and false negatives. Although a realistic value of the prevalence of deepfake videos may lead to a large $x=10^7$ (${\alpha=4\cdot 10^7}$), because this preview dataset has few true negatives, any false positives will lead to large variations in the wP metric, so we report wP values with $\alpha=100$ (note that this constant is subject to change for the final dataset). Since false positives are heavily weighted, wP is typically small, so we report $\log(\mathrm{wP})$ in our results; a value of 0 is the maximum achievable value, but generally the log-weighted-precision is negative.

\par Finally, although precision is paramount for very-rare detection problems (where detected content may undergo some form of human review or verification), recall must also be considered, as it is important to detect as many items of interest as possible. Therefore, we report the $\log(\mathrm{wP})$ for three levels of recall: $0.1$, $0.5$, and $0.9$. The weighted precision for each recall level can give a rough approximation of the cost of labeling videos if one wished to detect half, most, or nearly-all deepfakes in some real distribution of videos.
\section{Baseline}

\begin{table}
\begin{center}
\begin{tabular}{|l c c c |} 
\hline
\textbf{Method} & \textbf{Precision} & \textbf{Recall} & \textbf{log-WP}\\
\hline
TamperNet & 0.833 & 0.033 & -3.044\\
\hline
XceptionNet (Face) & 0.930 & 0.084 & -2.140\\
\hline
XceptionNet (Full) & 0.784 & 0.268 & -3.352\\
\hline
\end{tabular}
\end{center}
\caption{Video-level test metrics when optimizing for $\log(\mathrm{wP})$.}
\label{Table:video_baseline}
\end{table}

\begin{table}
\begin{center}
\begin{tabular}{|l c c c |} 
\hline
\textbf{Method} & \textbf{R=0.1} & \textbf{R=0.5} & \textbf{R=0.9}\\
\hline
TamperNet & -2.796 & -3.864 & -4.041\\
\hline
XceptionNet (Face) & -1.999 & -3.012 & -4.081\\
\hline
XceptionNet (Full) & -3.293 & -3.835 & -4.081\\
\hline
\end{tabular}
\end{center}
\caption{Video-level $\log(\mathrm{wP})$ for various recall values}
\label{Table:video_baseline}
\end{table}

\par To derive an initial baseline, we measured the performance of three simple detection models. The first model was a frame-based model which we denote as TamperNet. TamperNet is a small DNN (6 convolutional layers plus a 1 fully connected layer) trained to detect low-level image manipulations, such as cut-and-pasted objects or the addition of digital text to an image, and although it was not trained only on deepfake images, it performs well in identifying digitally-altered images in general (including face swaps). The other two models are the XceptionNet face detection and full-image models, trained on the FaceForensics data set \cite{roessler2019faceforensics}, and evaluated as implemented in \cite{ondyari2019}. For these models, one frame was sampled per second of video.


\par When using frame-based models for detection, there are two thresholds to tune - the per-frame detection threshold and a threshold that specifies how many frames must exceed the per-frame threshold in order to identify a video as fake (or the frames-per-video threshold). These thresholds must be tuned in tandem - for good performance, a low per-frame threshold will probably result in a high frames-per-video threshold, and vice-versa. To normalize for video length, we only evaluated the frames-per-video threshold over frames that contained a detectable face. During cross-validation on the train set, we found the optimal frame- and video-thresholds that maximized the log-WP over each fold, while still maintaining the desired level of recall. We then used these thresholds tuned on the training set to compute the metrics in Table~\ref{Table:video_baseline}.

\section{Closing remarks}
\par This document introduces a preview of the DFDC dataset that will be released later in the year with the intention to encourage researchers getting familiar with the data, provide early results and compare those with the proposed baselines. 

{\small
\bibliographystyle{IEEEtran}
\bibliography{main}
}

\end{document}